\PassOptionsToPackage{bookmarksnumbered,unicode}{hyperref}
\RequirePackage{hyperref}

\documentclass[sigconf]{aamas}

\makeatletter
\@ifpackageloaded{hyperref}{}{%
  \usepackage[hidelinks]{hyperref}
}
\makeatother

\usepackage{placeins} 
\usepackage{amsmath,amssymb}
\usepackage{graphicx}
\usepackage{subcaption}
\usepackage{booktabs}
\usepackage{multirow}
\usepackage{microtype}
\usepackage{balance}
\usepackage{enumitem}
\usepackage{siunitx}
\usepackage[dvipsnames]{xcolor}
\usepackage[most]{tcolorbox}
\usepackage{caption}
\usepackage{tabularx}
\usepackage{adjustbox}
\usepackage{tikz}
\usetikzlibrary{arrows.meta,positioning,fit,calc,shapes.geometric,shapes.misc,shapes.symbols}

\usetikzlibrary{arrows.meta,positioning,fit,calc,shapes.misc}
\usetikzlibrary{positioning,fit,calc,arrows.meta}
\tikzset{
  >={Latex[length=2.0mm]},
  box/.style       ={draw, rounded corners=2pt, minimum height=6.5mm, minimum width=26mm,
                     align=center, inner sep=2.5pt, very thick, fill=white},
  sbox/.style      ={box, minimum width=22mm},
  group/.style     ={draw, rounded corners=3pt, dashed, color=black!45, inner sep=5pt},
  arrow/.style     ={-Latex, line width=0.9pt},
  darrow/.style    ={-Latex, line width=0.6pt, dashed, color=black!55},
  note/.style      ={font=\footnotesize, color=black!65}
}

\setcopyright{ifaamas} \acmConference[AAMAS '26]{Proc.\@ of the 25th International Conference on Autonomous Agents and Multiagent Systems (AAMAS 2026)}{May 25-29, 2026} {Paphos, Cyprus}{Anonymous (eds.} 
\copyrightyear{2026} 
\acmYear{2026} 
\acmDOI{} 
\acmPrice{} 
\acmISBN{} 
\acmSubmissionID{697}
\title[Learning When to Ask]{Learning When to Ask: Simulation-Trained Humanoids for Mental-Health Diagnosis}

\author{Filippo Cenacchi}
\affiliation{\institution{Macquarie University}\city{Sydney}\country{Australia}}
\email{filippo.cenacchi@hdr.mq.edu.au}

\author{Deborah Richards}
\affiliation{\institution{Macquarie University}\city{Sydney}\country{Australia}}
\email{deborah.richards@mq.edu.au}

\author{Longbing Cao}
\affiliation{\institution{Macquarie University}\city{Sydney}\country{Australia}}
\email{longbing.cao@mq.edu.au}

\begin{abstract}
Testing humanoid robots with users is slow, causes wear, and limits iteration and diversity. Yet screening agents must learn to converse—timing, prosody, backchannels—and what to attend to in faces and speech for diagnosis of Depression and Post-Traumatic Stress Disorder (PTSD). Most simulators lack policy learning with nonverbal dynamics, and many controllers prioritise task accuracy while underweighting trust, pacing, and rapport. Virtualising the humanoid as a conversational agent in simulation offers a way to train models without hardware burden. We present an agent-centred, simulation-first pipeline that turns interview data into 276 Unreal Engine MetaHuman patients with synchronised speech, face/gaze, and head–torso poses, plus Patient Health Questionnaire–8 (PHQ-8) and PTSD Checklist—Civilian Version (PCL-C) flows. A perception–fusion–policy loop chooses what and when to speak, when to backchannel, and how to avoid interruptions, under a safety shield. Training uses counterfactual replay (bounded nonverbal perturbations) and an uncertainty-aware turn manager that targets probes to reduce diagnostic ambiguity. Results are simulation-only; the humanoid is the transfer target. Comparing three deep-learning models, our costumed TD3 (Twin Delayed (Deep Deterministic Gradient)) showed the largest improvement versus PPO (Proximal Policy Optimization) and CEM (Cross-Entropy Method), reaching near-ceiling coverage with higher pace stability at comparable final rewards. Decision-quality analyses indicated negligible turn overlap, aligned cut timing, fewer clarification prompts, and shorter waits. Performance remained stable under modality dropout and a renderer swap, and method ranking held on a held-out patient split. Contributions: (1) an agent-centred simulator that turns interviews into 276 interactive patients with bounded nonverbal counterfactuals; (2) a safe learning loop that treats timing and rapport as first-class control variables; (3) a comparative study (TD3 vs PPO/CEM) with clear gains in completeness and social timing; and (4) ablations and robustness analyses explaining why the gains arise, providing a reproducible path toward clinician-supervised humanoid pilots.
\end{abstract}

\keywords{Autonomous Agents, Multiagent Systems, Humanoid Robots, Simulation, Multimodal Diagnostics, Reinforcement Learning, Mental Health}

\begin{document}
\maketitle

\section{Introduction}

\begin{figure}[H]
  \centering
  \includegraphics[width=\columnwidth,clip,trim=20 30 20 25]{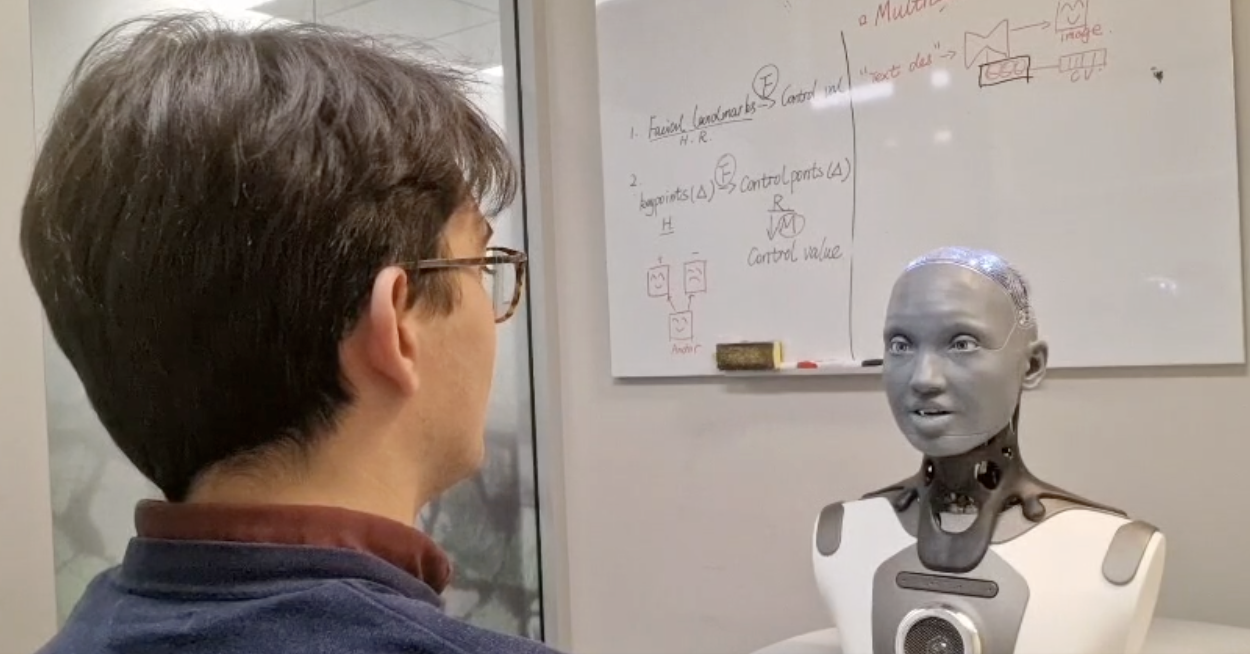}
  \vspace{-4pt}
  \Description{Ameca humanoid conversing with a user; repeated real-world use causes actuator wear.}
  \caption{Repeated real-world sessions cause actuators and mechanical wear to a humanoid robot.}
  \label{fig:ameca-burden}
\end{figure}

Humanoid agents are increasingly deployed as conversational partners in settings that demand sensitivity, reliability, and explicit safety constraints, including healthcare intake, psychotherapy adjuncts, eldercare coaching, and education support \cite{breazeal2016social,leite2013social}. Early clinical pilots indicate that robot- or agent-facilitated screening can preserve psychometric equivalence to clinician-led assessments while reducing stigma and facilitating disclosure, two preconditions for early detection of depressive and post-traumatic stress symptoms \cite{matsushima2024robot,di2019moca}. Robust deployment, however, hinges on mastering interactional competencies that are difficult to acquire within scarce, ethically constrained clinical windows, especially when diagnostic judgments depend on integrating lexical content with subtle nonverbal cues such as gaze aversion, flattened affect, and prosodic hesitation \cite{cummins2015review,ma2020multimodal}. Humanoid skill acquisition directly on hardware does not scale: each hour of user-facing training for platforms like Ameca incurs mechanical wear (neck/eye actuators, joint thermals), sensor recalibration, technician time, and room scheduling (Fig. ~\ref{fig:ameca-burden}). 

Clinical constraints further limit the number and diversity of participants. At population scale, this becomes prohibitive for improving speech recognition in noise, facial-cue interpretation, and adaptation to user behavior. \emph{We therefore virtualize training with realistic avatars}: controllable digital patients provide dense, repeatable experience and allow safe manipulation of gaze, affect, and timing signals, front-loading learning before any in-clinic exposure. Simulation has proven essential for scaling embodied skill learning without risking people or hardware \cite{tobin2017domain,dosovitskiy2017carla}. In human–robot interaction (HRI), controllable digital humans make it possible to manipulate gaze, head motion, timing, and prosody while preserving ecological validity for conversational studies \cite{jonell2017fusing,pan2008virtual}. Yet psychiatry-focused dialogue remains underrepresented in simulation pipelines: although interview corpora exist, they are often treated as offline datasets rather than transformed into \emph{interactive} patient populations that support policy learning under realistic multimodal dynamics \cite{gratch2014daic,ringwald2023edaic}. Additionally, many controllers prioritize task accuracy while underweighting trust, pacing, and rapport factors central to clinical acceptability and deployment \cite{wagner2023review}. We address this gap with a simulation framework that couples the Ameca humanoid to a pool of \textbf{276} interactive patients instantiated from E-DAIC (Extended Distress Analysis Interview Corpus) interviews \cite{gratch2014daic,ringwald2023edaic}. Our environment employs \textbf{Unreal Engine~5 MetaHumans} with high-fidelity facial animation, physically plausible eye gaze, and head–torso kinematics. Each patient exposes synchronized speech, facial action units, gaze vectors, and posture signals, and supports bounded counterfactual perturbations of nonverbal behavior during PHQ-8 and PCL-C dialogues. This substrate enables controlled “what-if” analyses of nonverbal cues while maintaining privacy by parameterizing behavior rather than copying identity features. 

Building on this simulator, we study how adaptive probing policies can improve triage quality under multimodal uncertainty. We compare two strong baselines \textbf{PPO} (Proximal Policy Optimization) \cite{schulman2017proximal} and a \textbf{CEM} (Cross-Entropy Method) style policy search with a \textbf{TD3} (Twin Delayed (Deep Deterministic Policy Gradient)) variant tailored to conversational decision-making with continuous rapport features and safety guardrails \cite{de2020surveyRLdialogue,wagner2023review}. We implement TD3 following the standard template~\cite{fujimoto2018td3} but adapt it to conversational control; throughout, we denote this instantiation as \emph{TD3 (ours)} and detail its departures from originally proposed TD3 in Sec.~\ref{sec:td3-vs-ppo-cem}. Policies are trained to balance diagnostic accuracy and class-wise sensitivity (Depression/PTSD) with interaction-quality metrics that capture pacing, turn-taking, and rapport. To accelerate experimentation and reduce clinician time, the framework includes replayable episodes, uncertainty-aware turn management, and batchable avatar cohorts, enabling rapid iteration on probe strategies before any in-clinic exposure.

\textbf{Contributions.} This paper makes four contributions:
(i) a Meta Human based, high-realism patient simulator that converts multimodal interview corpora into \emph{interactive} avatar populations with counterfactual nonverbal perturbations;
(ii) a reproducible pipeline for \emph{speeding up humanoid testing} via replay, uncertainty-aware control, and cohort batching that front-loads learning in simulation while enforcing HRI safety guardrails;
(iii) a comprehensive comparison of \textbf{PPO}, \textbf{CEM}, and a domain-tailored \textbf{TD3} controller on multimodal diagnostic probing over \textbf{276} patients, using accuracy, class sensitivities, rapport, and convergence speed as endpoints;
(iv) evidence that uncertainty-aware continuous control with counterfactual replay yields the largest gains from initialization in \emph{Coverage}, \emph{Rapport}, and \emph{Pace}, while maintaining near-ceiling \emph{Coverage} and strong decision quality; ablations and robustness analyses identify the key drivers.

\textbf{Paper structure.} Section~\ref{sec:related} reviews prior work on simulation for HRI, multimodal mental-health computing, and reinforcement learning (RL) for dialogue. Section~\ref{sec:system} describes the simulator and safety architecture; Section~\ref{sec:questionnaires} details PHQ-8 (Patient Health Questionnaire—8 item) and PCL-C (PTSD Checklist—Civilian Version) instrumentation; and Section~\ref{sec:td3-vs-ppo-cem} formalizes our objectives and TD3/PPO/CEM setups. Section~\ref{sec:results} reports empirical results, and Section~\ref{sec:ablations} presents ablations and robustness, followed by Discussion and Conclusion, in Sections \ref{sec:discussion} and \ref{sec:conclusion}, respectively.

\section{Related Work}
\label{sec:related}

\subsection{Simulation and High-Fidelity Avatars for Clinical HRI}
Simulation is a cornerstone for scaling embodied skill learning while protecting people and hardware \cite{tobin2017domain,dosovitskiy2017carla}. For conversational HRI, prior work shows that controllable digital humans let researchers manipulate gaze, prosody, timing, and head–torso kinematics with ecological validity \cite{pan2008virtual,jonell2017fusing}. In mental-health contexts, virtual interviewers have elicited sensitive disclosures and supported screening, pointing to stigma-reduction benefits and practical feasibility \cite{devault2014simsensei,lucas2014itsonlyacomputer,di2019moca,matsushima2024robot}. Recent pipelines provide the expressive control surface needed to study clinical micro-behaviors. \href{https://www.unrealengine.com/metahuman}{MetaHuman Creator} and \href{https://dev.epicgames.com/documentation/en-us/metahuman/metahuman-animator}{MetaHuman Animator} surface \emph{Facial Action Coding System} (FACS) / \emph{Apple ARKit} (ARKit) controls in \emph{Unreal Engine 5} (UE5); \href{https://developer.apple.com/documentation/arkit/arfacetrackingconfiguration}{ARKit Face Tracking} standardizes 52 blendshapes for cross-rig compatibility; and \href{https://docs.omniverse.nvidia.com/extensions/latest/ext_audio2face.html}{Omniverse Audio2Face} yields speech-synchronous visemes. Recent evidence underscores that human-likeness is multi-dimensional and bounded by perceptual/biological constraints \cite{swiechowski2025humanlike}, that embodiment increases social presence and enjoyment in older adults \cite{banck2025cards}, and that aligning verbal and gestural behaviors to personality improves communication satisfaction \cite{banna2025beyond}. In contrast to neutral \emph{GL Transmission Format Binary} (GLB) avatars, MetaHumans expose a clinically meaningful, frame-accurate control surface (FACS/ARKit blendshapes, gaze rays, head–neck chains) inside UE5, enabling controlled nonverbal perturbations and evaluation under realistic sensing and latency budgets conditions necessary for clinical HRI transfer \cite{jonell2017fusing,pan2008virtual}.

\subsection{Multimodal Mental-Health Computing and Clinical Corpora}
A robust literature links speech prosody, lexical/discourse markers, facial \emph{action units} (AUs) and gaze, and posture to depressive and PTSD symptomatology \cite{cummins2015review,ma2020multimodal}. The E-DAIC family provides synchronized audio–video–text interviews and clinical labels, catalyzing tri-modal benchmarks and robustness studies \cite{gratch2014daic,ringwald2023edaic,wagner2023review}. Tools such as OpenFace (AUs, gaze) and OpenPose (head/shoulder/torso) enable reliable feature extraction for research-grade HRI analysis \cite{baltrusaitis2018openface,cao2017openpose}. Beyond feasibility, studies of robot-mediated or virtual screening show comparable psychometrics and high acceptance when empathetic behaviors and guardrails are present \cite{di2019moca,matsushima2024robot,kumazaki2019brief}. Recent research strengthen two points our system operationalizes. First, multimodal fusion outperforms unimodal signals for Depression detection and is more robust to missing channels \cite{drougkas2024Depressionreview}. Second, combining \emph{large language models} (LLMs) with facial dynamics over interview-style data improves screening accuracy and interpretability, highlighting the incremental value of visual micro-cues over text alone \cite{npj2024multimodalDepression}. In line with guidance on trustworthy evaluation \cite{wagner2023review}, we fuse Whisper/ECAPA speech embeddings with OpenFace AU/gaze and OpenPose posture, track per-modality confidence, and apply bounded, clinically informed counterfactual perturbations (AUs, gaze, prosody) to stress-test policies against realistic variability rather than single-trajectory overfitting.

\subsection{Reinforcement Learning for Dialogue Probing and Trust-Aware Control}
 RL has been widely explored for dialogue management, including actor–critic and proximal objectives for stable policy updates \cite{schulman2017proximal}. Surveys from 2023–2024 document a move toward RL-enhanced controllers (and \emph{Reinforcement Learning from Human Feedback} (RLHF) variants) that optimize interaction-level metrics—not just slot/task accuracy—under uncertainty and partial observability \cite{kwan2023lmdialogsurvey,wang2024rlenhancedllms}. Clinical interviewing raises additional constraints: sparse rewards, long-horizon credit assignment, safety limits on admissible actions, and the need to encode rapport (latency alignment, interruption avoidance) alongside diagnostic performance \cite{de2020surveyRLdialogue,wagner2023review}. In this landscape, on-policy updates (e.g., PPO) remain strong baselines for stability \cite{schulman2017proximal}, while sampling-based search (e.g., CEM) is competitive for short-horizon static tuning; deterministic off-policy controllers are often preferred when actions are smooth and bounded (e.g., timing/immediacy controls) and when replay can be exploited. Consistent with these trends, our work focuses on uncertainty-aware multimodal encoding \cite{sun2020adversarialmultimodal}, counterfactual regularization over nonverbal cues, and a rule-based safety layer aligned with socially assistive robotics and AI ethics guidance \cite{nakamura2022safetyrisk,mittelstadt2016ethics}.

\section{System Overview}
\label{sec:system}

\begin{figure}[t]
  \centering
  \includegraphics[width=\columnwidth]{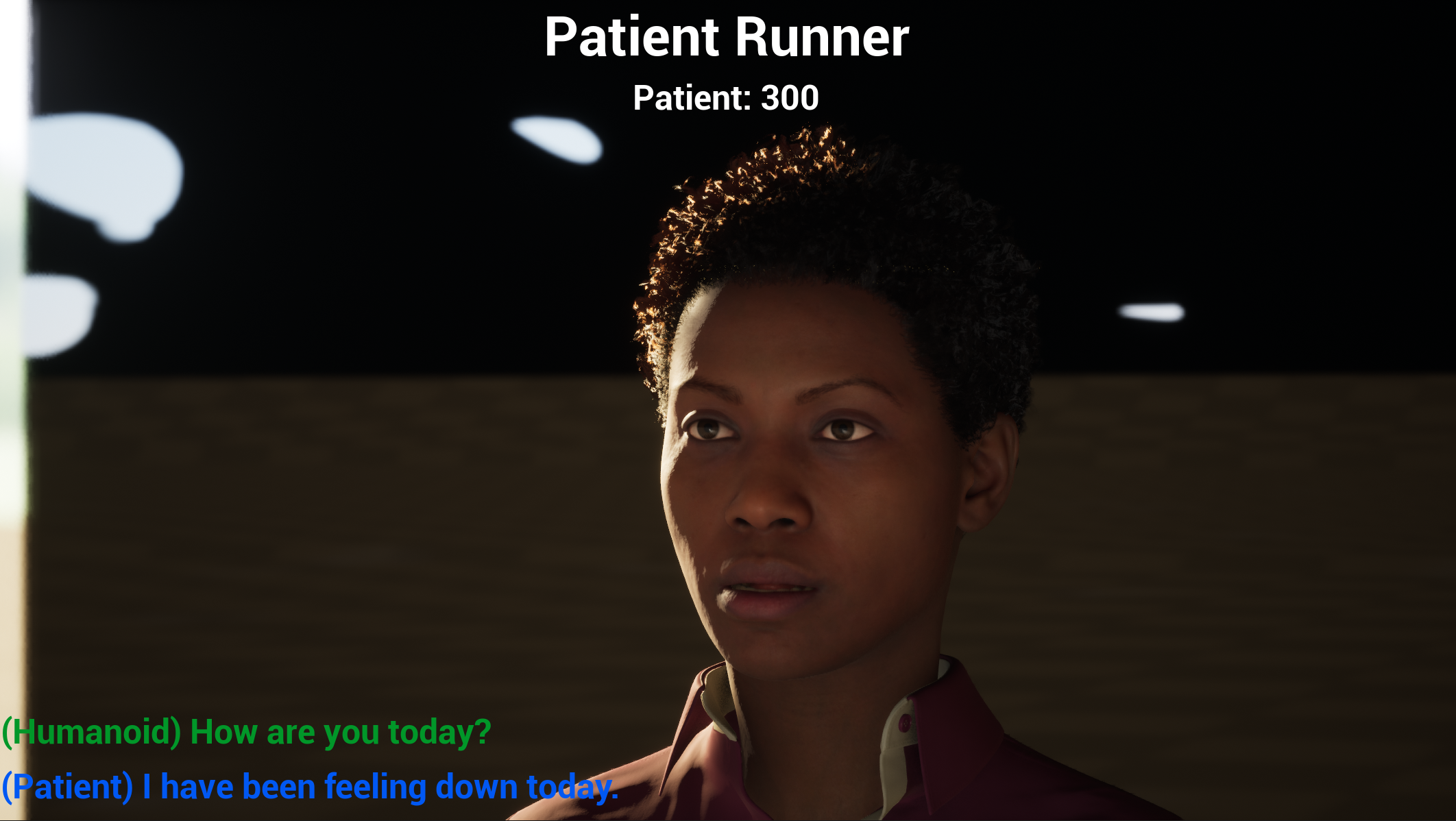}
  \caption{In-sim screenshot (UE5 MetaHuman patient runner). The pipeline uses AI-powered MetaHumans, enabling realistic facial animation for conversational diagnostics.}
  \Description{UE5 MetaHuman patient runner screenshot used to simulate clinical interviews.}
  \label{fig:scene}
  \vspace{-0.2em} 
\end{figure}

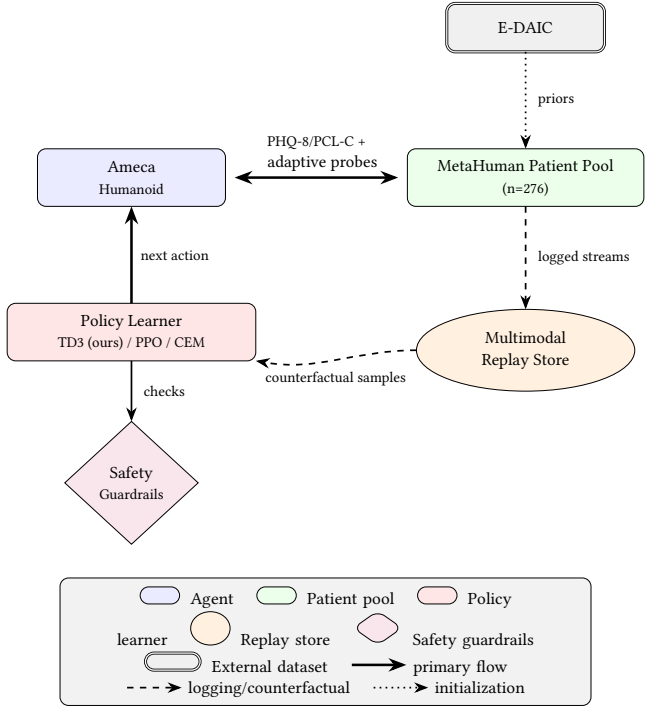
\begin{figure}[t]
\centering
\makebox[\columnwidth][c]{
\adjustbox{width=\columnwidth}{%
\begin{tikzpicture}[>=Stealth, font=\small, node distance=1.0cm, every node/.style={outer sep=0pt}]
\tikzset{
  agent/.style={rounded corners, draw, fill=blue!8, minimum width=3.1cm, minimum height=0.95cm, align=center},
  pool/.style={rounded corners, draw, fill=green!8, minimum width=3.9cm, minimum height=0.95cm, align=center},
  learner/.style={rounded corners, draw, fill=red!10, minimum width=4.1cm, minimum height=0.95cm, align=center},
  store/.style={ellipse, draw, fill=orange!12, minimum width=3.6cm, minimum height=1.35cm, align=center},
  guard/.style={diamond, draw, fill=purple!10, minimum width=2.2cm, align=center},
  dataset/.style={double, rounded corners, draw, fill=gray!12, minimum width=2.6cm, minimum height=0.8cm, align=center},
  legendbox/.style={draw, rounded corners, fill=gray!10, inner sep=4pt, align=left},
  legendicon/.style={minimum width=0.65cm, minimum height=0.28cm}
}

\node[agent]   (hum)  {Ameca\\{\footnotesize Humanoid}};
\node[pool, right=3.0cm of hum] (pool) {MetaHuman Patient Pool\\{\footnotesize (n=276)}};
\node[store, below=1.7cm of pool] (replay) {Multimodal\\Replay Store};
\node[learner, below=1.6cm of hum] (learner) {Policy Learner\\{\footnotesize TD3 (ours) / PPO / CEM}};
\node[guard,  below=1.0cm of learner] (safety) {Safety\\{\footnotesize Guardrails}};
\node[dataset, above=1.6cm of pool] (data) {E-DAIC};

\draw[<->, very thick, shorten <=1.5mm, shorten >=1.5mm]
  (hum.east) -- node[above, xshift=1mm]
  {\shortstack[l]{\footnotesize PHQ-8/PCL-C +\\ adaptive probes}}
  (pool.west);

\draw[->, dashed, thick] (pool.south) -- node[right, xshift=1mm]{\footnotesize logged streams} (replay.north);
\draw[->, dashed, thick] (replay.west) to[out=180,in=350] node[below, yshift=-1mm]{\footnotesize counterfactual samples} (learner.south east);
\draw[->, very thick] (learner.north) -- node[right, xshift=0.5mm]{\footnotesize next action} (hum.south);
\draw[->, dotted, thick] (data.south) -- node[right, xshift=1mm]{\footnotesize priors} (pool.north);
\draw[->, thick] (learner) -- node[right, xshift=1mm]{\footnotesize checks} (safety);

\node[legendbox, anchor=north, text width=\linewidth, align=center]
  at ([yshift=-0.55cm]current bounding box.south)
  {%
   \tikz{\node[agent,   legendicon] {};}\; Agent \quad
   \tikz{\node[pool,    legendicon] {};}\; Patient pool \quad
   \tikz{\node[learner, legendicon] {};}\; Policy learner \quad
   \tikz{\node[store,   legendicon, minimum height=0.55cm] {};}\; Replay store \quad
   \tikz{\node[guard,   legendicon, minimum width=0.8cm, minimum height=0.5cm] {};}\; Safety guardrails \quad
   \tikz{\node[dataset, legendicon, minimum width=0.9cm] {};}\; External dataset \quad
   \tikz{\draw[->, very thick] (0,0) -- (0.9,0);} primary flow \quad
   \tikz{\draw[->, dashed, thick] (0,0) -- (0.9,0);} logging/counterfactual \quad
   \tikz{\draw[->, dotted, thick] (0,0) -- (0.9,0);} initialization
  };

\end{tikzpicture}}}
\caption{Closed loop for humanoid training with MetaHuman patients.}
\label{fig:pipeline}
\end{figure}

Our system trains a simulated Ameca digital twin, through conversations with a cohort of 276 MetaHuman patients rendered in Unreal Engine 5 (UE5) (see Fig.~\ref{fig:scene}). Each avatar encapsulates: (i) a PHQ-8 and PCL-C questionnaire state machine; (ii) synchronized multimodal generators—speech, facial action units (AUs) and gaze (OpenFace), and head–shoulder–torso pose (OpenPose); and (iii) clinically bounded perturbation ranges learned from E-DAIC statistics. The closed loop in Fig.~\ref{fig:pipeline} alternates mandatory items with adaptive probes proposed by the learning policy and vetted by a safety layer. All audio/text/AU/gaze/pose/timing streams are logged to a replay store for counterfactual sampling and cohort batching. Throughout the whole experiment, all results are from the simulation; the physical Ameca is only the transfer target and informs sensing/latency constraints.

\paragraph{Humanoid–Avatar Interaction Loop.}
At each turn $t$, the policy receives rendered speech and video. We extract: (a) Whisper/ECAPA (Emphasized Channel Attention, Propagation and Aggregation) speech embeddings \cite{radford2023whisper,desplanques2020ecapa}; (b) OpenFace AUs and gaze rays \cite{baltrusaitis2018openface}; and (c) OpenPose head/shoulder/torso pose \cite{cao2017openpose}. We also compute per-modality reliability scores $\kappa_t^{(m)}$ from extractor diagnostics (e.g., \emph{ASR} (Automatic Speech Recognition) proxy \emph{WER} (Word Error Rate), OpenFace confidence) so unreliable channels are down-weighted \cite{ma2020multimodal,wagner2023review}. A turn manager enforces minimum/maximum dwell, inserts neutral immediacy behaviors (backchannels, nods) when policy entropy is high, and regulates pace/overlap to preserve rapport \cite{jonell2017fusing}. This corresponds to the thick bidirectional arrow in Fig.~\ref{fig:pipeline}; dashed arrows depict logging to the replay store.

\paragraph{MetaHuman Patient Runner.}
Each patient is a UE5 MetaHuman with a \emph{FACS} (Facial Action Coding System)/\emph{Apple ARKit} (ARKit) blendshape rig, controllable eye-gaze rays, and a head–neck chain (Fig.~\ref{fig:scene}). To study cue sensitivity without reproducing identity, we \emph{parameterize behavior} (AU intensities, fixation maps, head tilt, shoulder slump) rather than copying raw appearance/voice \cite{buxbaum2018privacyavatars,voas2017digitaltwin}. For “what-if’’ analyses, we create \emph{counterfactual} versions of a turn by applying small, clinically plausible changes to nonverbal cues (e.g., slightly stronger AU4 brow-lowerer or more gaze aversion) learned from E-DAIC distributions \cite{ringwald2023edaic,gratch2014daic}; we reject biomechanically implausible poses to preserve ecological validity.

\begin{figure}[t]
\centering
\adjustbox{max width=\columnwidth}{
\begin{tikzpicture}[>=Stealth, font=\small, node distance=0.4cm]
\tikzset{
  enc/.style={draw, rounded corners, fill=blue!10, minimum width=3.1cm, minimum height=0.9cm, align=center},
  fuse/.style={draw, rounded corners, fill=orange!15, minimum width=4.6cm, minimum height=2.2cm, align=center},
  heads/.style={draw, rounded corners, fill=red!10, minimum width=3.6cm, minimum height=0.9cm, align=center},
  store/.style={cylinder, draw, fill=green!12, aspect=1.4,
              minimum height=1.0cm, minimum width=1.8cm, align=center},
  legendbox/.style={draw, rounded corners, fill=gray!10, inner sep=4pt, align=left},
  legendicon/.style={minimum width=0.65cm, minimum height=0.28cm}
}

\node[enc] (sp) {Speech Encoder\\{\footnotesize Whisper + ECAPA}};
\node[enc, below=0.7cm of sp] (fa) {Face/Gaze Encoder\\{\footnotesize OpenFace (AUs, gaze)}};
\node[enc, below=0.7cm of fa] (po) {Pose Encoder\\{\footnotesize OpenPose (head/shoulder/torso)}};

\node[fuse, right=1.4cm of fa] (attn) {Transformer Fusion\\{\footnotesize cross-modal + uncertainty}};
\node[heads, right=1.6cm of attn] (hv) {Policy/Value Heads\\{\footnotesize TD3 (ours) / PPO / CEM}};
\node[store, below=1.1cm of attn] (cf) {Counterfactual\\Replay Buffer};

\draw[->, very thick] (sp) -- (attn);
\draw[->, very thick] (fa) -- (attn);
\draw[->, very thick] (po) -- (attn);
\draw[->, very thick] (attn) -- (hv);
\draw[->, dashed, thick] (cf) -- node[right]{\footnotesize plausible variants $s'_t$} (attn);

\node[legendbox, anchor=north, yshift=-0.55cm]
  at (current bounding box.south)
  {%
   \tikz{\node[enc,   legendicon] {};}\; Encoder \quad
   \tikz{\node[fuse,  legendicon] {};}\; Transformer fusion \quad
   \tikz{\node[heads, legendicon] {};}\; Policy/Value heads \quad
   \tikz{\node[store, legendicon, minimum height=0.55cm] {};}\; Counterfactual buffer \quad
   \tikz{\draw[->, very thick] (0,0) -- (0.8,0);} primary flow \quad
   \tikz{\draw[->, dashed, thick] (0,0) -- (0.8,0);} counterfactual variants
  };

\end{tikzpicture}}
\caption{Learning stack shared by PPO, CEM, and our TD3 variant. Encoders feed a transformer fusion block; heads and updates differ by algorithm.}
\label{fig:policy}
\end{figure}
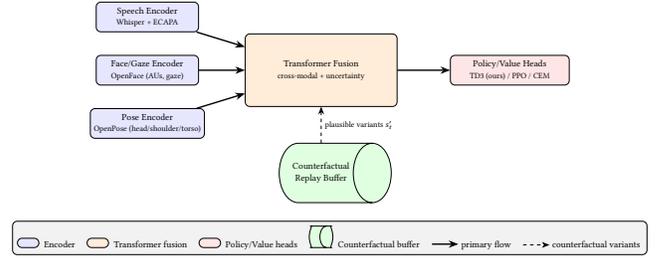

\paragraph{Policy Learning Stack.}
Modality-specific encoders map short windows of speech, face/gaze, and pose into fixed-length tokens; a transformer \emph{fusion} block conditions decisions on cross-modal context and the reliability scores (Fig.~\ref{fig:policy}) \cite{sun2020adversarialmultimodal,ma2020multimodal}. We compare three learners that share the encoders/fusion but differ in heads/updates: \emph{PPO} (Proximal Policy Optimization) as a stable on-policy baseline \cite{schulman2017proximal}; a sampling-based \emph{CEM} (Cross-Entropy Method) policy search; and a \emph{custom twin-critic off-policy variant} designed for smooth, bounded rapport controls (latency alignment, interruption penalties). We use a bounded continuous controller with \emph{two} action-value (critic) networks and a \emph{delayed} policy (actor) update, tuned for smooth rapport controls (latency alignment, interruption avoidance) under replay. The actor maps the fused representation to a 5-D action vector (timing/backchannel parameters) passed through a \emph{Sigmoid} and \emph{affine} scaling to the physical bounds $[\ell,h]$ used in our simulator (cf.\ Sec.~\ref{sec:td3-vs-ppo-cem}). To reduce spurious edge effects, we add small zero-mean Gaussian exploration noise during data collection and \emph{clamp} actions to $[\ell,h]$. Two critics $Q_{\phi_1},Q_{\phi_2}$ regress the return for $(s,a)$, and the target for bootstrapping uses the \emph{minimum} of the target critics evaluated on a \emph{smoothed} target action $\tilde a'$ (policy output plus clipped noise) to curb over-estimation near bounds:
\[
\tilde a'=\mathrm{clip}\!\big(\pi_{\theta'}(s')+\tilde\varepsilon,\ \ell,\ h\big),\qquad
y=r+\gamma\,\min\nolimits_{i\in\{1,2\}} Q_{\phi_i'}\!\big(s',\tilde a'\big).
\]
Critics minimize $\big(Q_{\phi_i}(s,a)-y\big)^2$. The actor maximizes $Q_{\phi_1}(s,\pi_{\theta}(s))$ but is updated on a slower cadence (policy delay) than the critics to stabilize learning under off-policy replay.

Two domain-specific regularizers make the controller robust to conversational variability. First, \emph{counterfactual consistency}: for states $s_t$ and their clinically plausible nonverbal variants $s_t'$ (small changes in \emph{action units} (AUs), gaze, pose, or prosody drawn from our counterfactual buffer), we penalize action drift:
\[
\mathcal{L}_{\text{cf}}=\lambda_{\text{cf}}\;\big\|\pi_\theta(s_t)-\pi_\theta(s_t')\big\|_2^2,
\]
encouraging stable timing decisions under realistic cue perturbations. Second, \emph{reliability-aware weighting}: the fused representation includes the per-modality reliability scalars $\kappa^{(m)}$; during training we stochastically drop low-$\kappa$ channels and rescale the fusion features so the policy learns to rely on whichever signals are trustworthy \cite{ma2020multimodal}. Targets use \emph{Polyak} averaging with coefficient $\tau$ for smooth parameter tracking. All components share the same encoders and transformer fusion shown in Fig.~\ref{fig:policy}; full hyperparameters and bounds appear in Sec.~\ref{sec:td3-vs-ppo-cem}.

\paragraph{Trust- and Uncertainty-Aware Control.}
The reward balances diagnostic progress and interaction quality:
\[
R=\alpha\,\Delta\mathrm{Acc}+\gamma\,\mathrm{Sens}_{\{\mathrm{Dep},\mathrm{PTSD}\}}+\rho\,\mathrm{Rapport},
\]
where \emph{Rapport} aggregates latency matching and interruption penalties \cite{wagner2023review,jonell2017fusing}. We train with \emph{counterfactual replay}: for some states $s_t$ we sample plausible variants $s'_t$ (tweaked AUs/gaze/pose/prosody) and regularize the policy toward consistent actions across $s_t\!\to\!s'_t$, improving robustness to realistic nonverbal variability \cite{sun2020adversarialmultimodal} (see the dashed arc from the replay store to the learner in Fig.~\ref{fig:pipeline}).

\paragraph{Safety and Auditability.}
A rule layer checks proposed probes against (i) whitelists and de-escalation templates, (ii) topic-wise dwell caps, and (iii) timeouts/fallbacks that favor well-being and oversight; every override is logged with the relevant reward terms to produce an auditable trace aligned with socially assistive robotics and AI ethics guidance \cite{nakamura2022safetyrisk,mittelstadt2016ethics} (Safety diamond in Fig.~\ref{fig:pipeline}).

\subsection{Clinical Questionnaires and Scoring}
\label{sec:questionnaires}
We operationalize standardized screeners for Depression and PTSD the \textbf{PHQ-8} and \textbf{PCL-C} as in-simulation sub-tasks that both structure the dialogue and supervise learning (Tables~\ref{tab:phq8}–\ref{tab:pclc}). In each episode, \emph{Ameca}, acting as the clinician agent, asks the mandatory PHQ-8 items (Table~\ref{tab:phq8}) and PCL-C items (Table~\ref{tab:pclc}); the patient agent replies in natural language (speech/text). A lightweight LLaMA (Large Language Model Meta AI) based interpreter maps each answer to the corresponding Likert score (0–3 for PHQ-8; 1–5 for PCL-C), with totals and severities computed per the scoring summaries in Tables~\ref{tab:phq8} and \ref{tab:pclc}. These scores serve (i) as the diagnostic signals for depressive symptoms and probable PTSD (cutpoint/cluster rules), and (ii) as control inputs that steer the interview: they trigger targeted follow-ups under high uncertainty or severity, shape reward terms (uncertainty reduction, class sensitivity), and parameterize avatar affect, gaze, and posture for counterfactual analyses. In parallel, the system synchronously logs facial action cues and speech prosody linked to each item as structured clinical notes, alongside the item-level trajectories.

\begin{table}[t]
\centering
\setlength{\tabcolsep}{6pt}
\renewcommand{\arraystretch}{1.08}
\caption{PHQ-8 items (0–3 Likert).}
\label{tab:phq8}
\small
\begin{tabular}{@{}p{2.2em}p{\dimexpr\linewidth-2.2em-2\tabcolsep\relax}@{}}
\toprule
\textbf{No.} & \textbf{Questions} \\
\midrule
Q1 & Little interest or pleasure in doing things \\
Q2 & Feeling down, depressed, or hopeless \\
Q3 & Trouble falling or staying asleep, or sleeping too much \\
Q4 & Feeling tired or having little energy \\
Q5 & Poor appetite or overeating \\
Q6 & Feeling bad about yourself—or that you are a failure or have let yourself or your family down \\
Q7 & Trouble concentrating on things, such as reading or watching TV \\
Q8 & Moving or speaking noticeably slowly—or being fidgety/restless more than usual \\
\bottomrule
\end{tabular}

\vspace{4pt}
\scriptsize
\textbf{Scoring.} 0=\emph{not at all}, 1=\emph{several days}, 2=\emph{more than half the days}, 3=\emph{nearly every day}. 
Sum 0–24; cutpoints \emph{5, 10, 15, 20} map to mild, moderate, moderately severe, severe; $\geq 10$ indicates probable current major Depression~\cite{kroenke2009phq8}.
\end{table}

\begin{table}[t]
\centering
\setlength{\tabcolsep}{6pt}
\renewcommand{\arraystretch}{1.08}
\caption{PCL-C items (1–5 Likert).}
\label{tab:pclc}
\small
\begin{tabular}{@{}p{2.2em}p{\dimexpr\linewidth-2.2em-2\tabcolsep\relax}@{}}
\toprule
\textbf{No.} & \textbf{Questions} \\
\midrule
Q1  & Repeated, disturbing memories, thoughts, or images of a stressful experience from the past? \\
Q2  & Repeated, disturbing dreams of a stressful experience from the past? \\
Q3  & Suddenly acting or feeling as if a stressful experience were happening again (as if reliving it)? \\
Q4  & Feeling very upset when something reminded you of a stressful experience from the past? \\
Q5  & Having physical reactions (e.g., heart pounding, trouble breathing, sweating) when reminded? \\
Q6  & Avoid thinking about or talking about a stressful experience or avoid having related feelings? \\
Q7  & Avoid activities or situations because they reminded you of a stressful experience from the past? \\
Q8  & Trouble remembering important parts of a stressful experience from the past? \\
Q9  & Loss of interest in things you used to enjoy? \\
Q10 & Feeling distant or cut off from other people? \\
Q11 & Feeling emotionally numb or unable to have loving feelings for those close to you? \\
Q12 & Feeling as if your future somehow will be cut short? \\
Q13 & Trouble falling or staying asleep? \\
Q14 & Feeling irritable or having angry outbursts? \\
Q15 & Having difficulty concentrating? \\
Q16 & Being “super-alert” or watchful on guard? \\
Q17 & Feeling jumpy or easily startled? \\
\bottomrule
\end{tabular}

\vspace{4pt}
\scriptsize
\textbf{Scoring.} 1=\emph{not at all} to 5=\emph{extremely}. Total 17–85;
clusters: B=Q1–Q5, C=Q6–Q12, D=Q13–Q17.
Two standards: (i) \emph{symptom-cluster} rule meeting B(1)+C(3)+D(2);
(ii) \emph{cutpoint} on total, commonly \emph{44–50} (\emph{44} for civilian screening)~\cite{blanchard1996pcl,ruggiero2003pcl}.
\end{table}

\paragraph{Integration with Dialogue and Reward.}
Mandatory questionnaire items are asked verbatim before policy-generated follow-ups. The policy’s \emph{uncertainty-aware probes} target items or clusters with highest posterior uncertainty, while rewards include (i) improvement in PHQ-8/PCL-C predictive certainty, (ii) class-wise sensitivity for Depression/PTSD screens, and (iii) rapport metrics (latency alignment, overlap penalties) \cite{wagner2023review,jonell2017fusing}. Counterfactuals perturb nonverbal cues during item delivery (e.g., varying AU12/AU4 intensity or gaze aversion) to quantify their causal influence on policy decisions.

\subsection{TD3 for Multi-Metric Interview Control, and Comparison to PPO and CEM}
\label{sec:td3-vs-ppo-cem}

\paragraph{Setting.}
Each episode is a 25-turn clinical-style interview. At every step $t$, the simulator emits a 10-D metric vector $m_t\!\in\!\mathbb{R}^{10}$ summarizing conversation progress and quality (e.g., \emph{coverage} of mandatory items, \emph{rapport} from latency/overlap composites, \emph{balance} of topic spread, \emph{pace} alignment, plus error/quality proxies). The agent’s observable state is a 20-D vector
\[
s_t \;=\; \big[x_t \,\|\, w\big],
\]
where $x_t\!\in\!\mathbb{R}^{10}$ are turn-level interaction features and $w\!\in\!\mathbb{R}^{10}$ encodes reviewer/clinician preferences over the same metrics. The action is a 5-D continuous control $a_t\!\in\!\mathbb{R}^{5}$ that parameterizes timing and backchannel behavior. Concretely, the five dimensions govern: (1) target response latency, (2) maximum wait before intervening, (3) backchannel rate, (4) interruption tolerance/threshold, and (5) a gain that scales immediacy-related micro-behaviors. The instantaneous reward is a linear scalarization of the metrics by the preferences,
\[
r_t \;=\; \langle w,\,m_t\rangle,
\]
and the objective is the discounted return with factor $\gamma$. This formulation lets us pose multi-objective conversational control as a single continuous-action decision problem while making the role of $w$ explicit: different reviewers can emphasize different trade-offs without changing the environment dynamics.

\paragraph{TD3 architecture (ours).}
We instantiate TD3 with design choices tailored to smooth, bounded rapport controls. Let $[\ell,h]$ denote per-dimension physical bounds used by the simulator:
\[
[\ell,h] \;=\; \big([10,\,3,\,0.40,\,0.00,\,0.85],\;[24,\,9,\,0.85,\,0.70,\,1.15]\big).
\]
\textbf{Actor} $\pi_\theta$: \texttt{Dense(256)+LayerNorm+SiLU (Sigmoid Linear Unit)} $\rightarrow$ \texttt{Dense(256)+SiLU} $\rightarrow$ \texttt{Dense(5)} $\rightarrow$ \texttt{Sigmoid} then affine scaling to $[\ell,h]$. During data collection we add zero-mean Gaussian exploration noise $\varepsilon\!\sim\!\mathcal{N}(0,0.06^2)$ and clip to $[\ell,h]$ so actions always respect safety/comfort limits. \textbf{Twin critics} $Q_{\phi_1},Q_{\phi_2}$ each take $[s,a]\!\in\!\mathbb{R}^{25}$ and use \texttt{Dense(256)+SiLU} $\rightarrow$ \texttt{Dense(256)+SiLU} $\rightarrow$ \texttt{Dense(1)}. Target networks track the online parameters via Polyak averaging with $\tau{=}0.005$. We apply target-policy smoothing by adding $\tilde\varepsilon\!\sim\!\mathcal{N}(0,0.04^2)$ (clipped to $\pm 0.08$) to the target action before clipping to $[\ell,h]$; this reduces value overestimation near action bounds and encourages locally coherent policies.

\paragraph{Learning rule.}
Given a minibatch of transitions, we compute a smoothed target action $\tilde a'=\mathrm{clip}\!\big(\pi_{\theta'}(s')+\tilde\varepsilon,\,\ell,\,h\big)$ and the TD3 target
\[
y \;=\; r \;+\; \gamma\,\min\nolimits_{i\in\{1,2\}} Q_{\phi_i'}\!\big(s',\,\tilde a'\big).
\]
Each critic minimizes $\big(Q_{\phi_i}(s,a)-y\big)^2$. The actor maximizes $Q_{\phi_1} (s,\pi_\theta(s))$ but is updated on a slower cadence than the critics (\emph{policy delay} of $2$), which empirically improves stability for noisy, partially observed conversational dynamics by letting value estimates settle before moving the policy.

\paragraph{Training pipeline.}
We store $(s,a,r,s')$ in a replay buffer of capacity $200{,}000$ and train with minibatches of size $256$. Both actor and critics use Adam with learning rate $3{\times}10^{-4}$; the discount is $\gamma{=}0.985$. Each episode lasts $25$ steps (covering the 8 PHQ-8 turns and 17 PCL-C turns described in Sec.~\ref{sec:questionnaires}). Unless otherwise noted, curves reported in Sec.~\ref{sec:results} are means across random seeds with a 35-step rolling window to smooth short-term variance while preserving learning trends. This setup makes efficient use of expensive simulated conversations (via replay) and yields reproducible learning curves suitable for ablation and policy comparisons.

\paragraph{Why this design fits the domain.}
(i) \emph{Bounded outputs} (Sigmoid$+\!$ affine scaling) keep timing/intensity within clinically acceptable ranges without ad-hoc clamps at inference. (ii) \emph{SiLU} activations support fine-grained, non-saturating adjustments—useful when nudging latency or backchannel frequency rather than making abrupt changes. (iii) \emph{Twin critics + policy delay} curb value overestimation and stabilize off-policy updates in the presence of stochastic user behavior and partial observability. (iv) \emph{Target-policy noise} improves target value estimates near bounds precisely where conversational parameters often reside due to safety/comfort limits.

\paragraph{Baselines.}
\textbf{PPO (Proximal Policy Optimization)} \cite{schulman2017proximal} uses a single actor–critic \emph{MLP (Multi-Layer Perceptron)} with \emph{SiLU}, the clipped-ratio objective, \emph{GAE (Generalized Advantage Estimation)} with $\lambda{=}0.92$, clip range $\pm 0.2$, an entropy coefficient of $0.004$, and a standard value loss. PPO is a robust on-policy baseline that steadily improves \emph{coverage} and \emph{rapport}, but by construction discards most data and therefore adapts more slowly than off-policy TD3 in our simulator; late training often reveals a mild \emph{pace} slowdown as the policy converges to a single preferred cadence. \textbf{CEM (Cross-Entropy Method)} treats the 5-D action as a population-optimized parameter: we sample $64$ candidates, keep the top $25\%$ as elites, and update the mean/variance. CEM can quickly find high-performing static settings on short horizons, but with no state feedback or credit assignment its improvements (\emph{deltas}) diminish as horizon and dimensionality grow; empirically it becomes sample-hungry compared to TD3/PPO for our sequential control task.

\paragraph{Empirical summary.}
Across runs (see Sec.~\ref{sec:results}) all methods converge to a narrow band of final reward, but TD3 and PPO achieve the \emph{largest improvements from initialization}. TD3 yields the biggest gains in \emph{Coverage} (approaching a ceiling), \emph{Rapport} (driven by lower overlap and better latency alignment), and \emph{Pace} (faster, more consistent turn timing). CEM often starts strong (good initial \emph{Coverage}) but exhibits very small subsequent \emph{deltas}. \emph{Balance} (probe/topic spread) changes little across policies, with PPO edging upward slightly more than TD3/CEM. Decision-quality endpoints show that our TD3 controller attains zero overlap with perfect cut consistency while maintaining near-ceiling \emph{Coverage}; error proxies (e.g., unnecessary waits/clarifications) converge to similarly low levels across methods.

\paragraph{Key hyperparameters.}
Discount $0.985$; Polyak $\tau{=}0.005$; Adam learning rate $3{\times}10^{-4}$ (actor and critics); replay capacity $200{,}000$; batch size $256$; exploration noise $\sigma{=}0.06$; target-policy noise $\sigma{=}0.04$ (clip $\pm 0.08$); policy delay $=2$; actor head \texttt{Sigmoid} $\to$ scale to $[\ell,h]$ (bounds as above). PPO: $\lambda{=}0.92$, clip $\pm 0.2$, entropy $0.004$. CEM: population $64$, elite fraction $25\%$.

\section{Results and Analysis}
\label{sec:results}

We evaluate \textbf{PPO}, \textbf{CEM}, and \textbf{our TD3 variant} at cohort scale over \textbf{276} MetaHuman patients, reporting both \emph{learning dynamics} and \emph{end-of-training} outcomes. We track five application-facing metrics—overall episodic \emph{Reward}, interview \emph{Coverage} (completeness of required items), conversational \emph{Rapport} (latency/overlap composite), \emph{Balance} (topic spread and probe diversity), and \emph{Pace} (turn-timing alignment)—chosen to articulate autonomous-agent performance beyond task accuracy and into interaction quality.

\begin{figure}[t]
  \centering
  \includegraphics[width=\columnwidth]{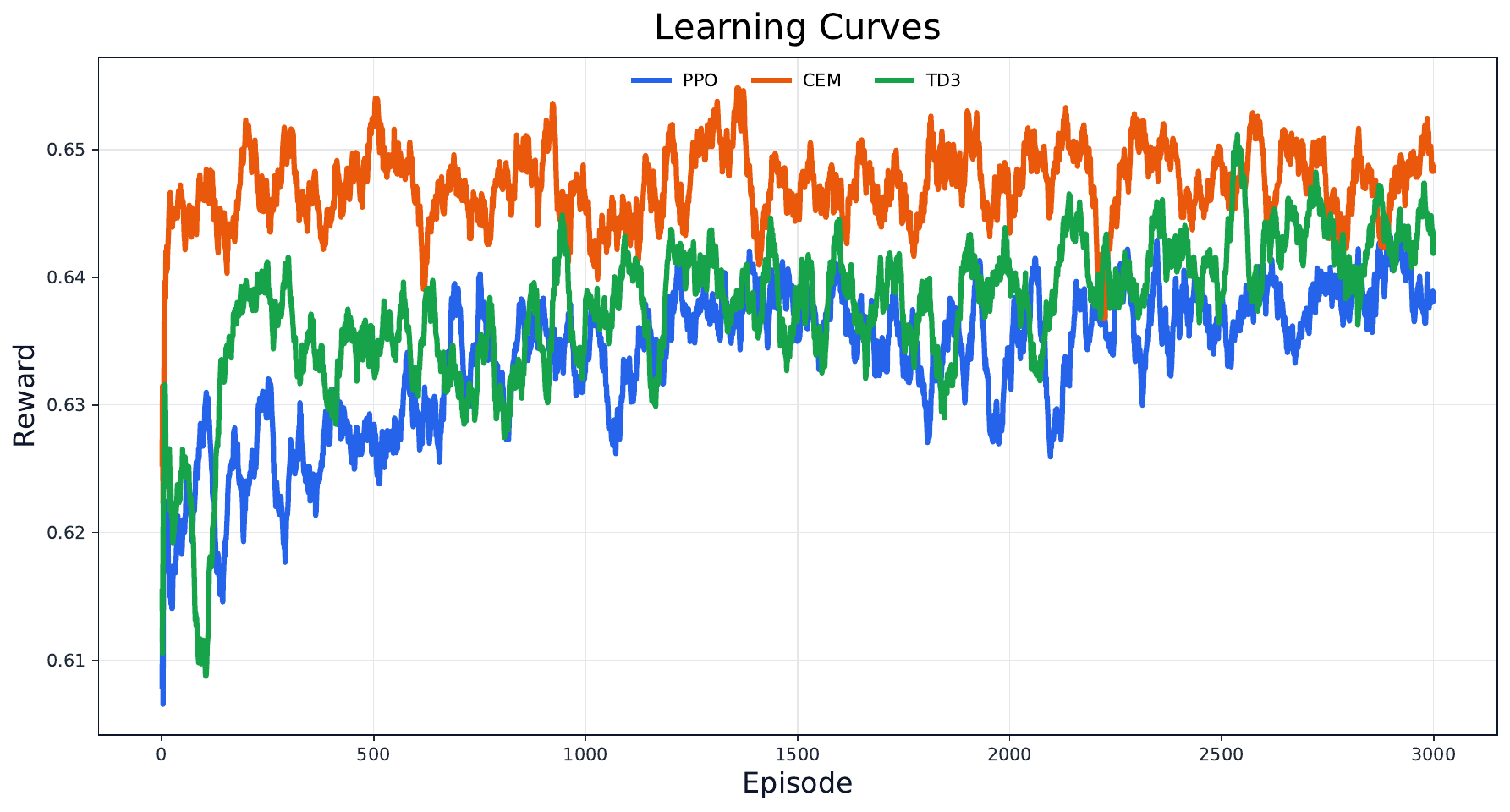}
  \caption{Learning curves for episodic reward (mean across runs). All methods improve across training and converge within a narrow band.}
  \label{fig:learn}
\end{figure}

\begin{figure}[t]
  \centering
  \includegraphics[width=\columnwidth]{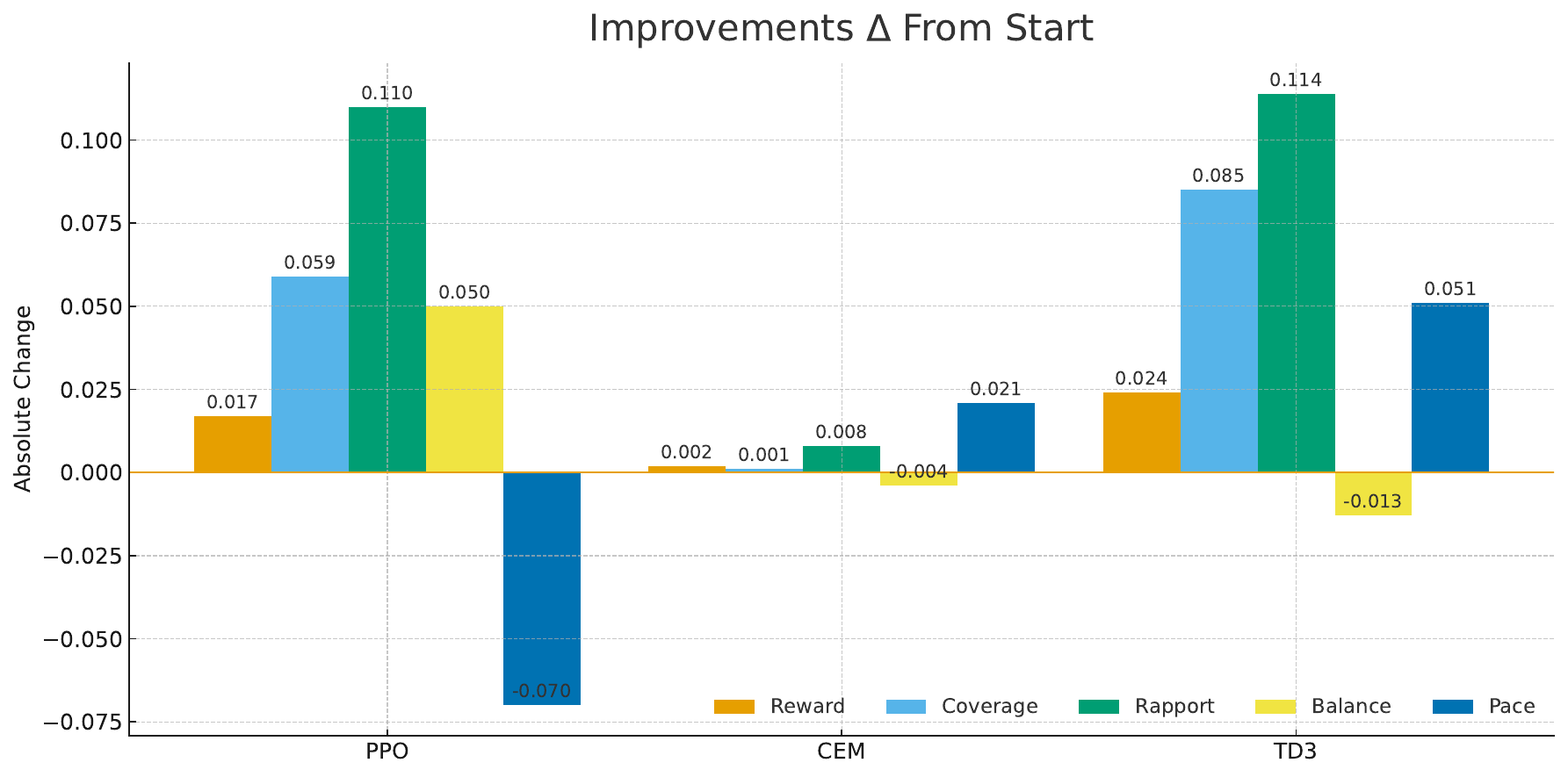}
  \caption{\textbf{Absolute improvement ($\Delta$) from initialization.} Bars report gains by metric; higher is better.}
  \label{fig:delta}
\end{figure}

\paragraph{Learning dynamics.} Figure~\ref{fig:learn} charts reward over 3k episodes. All policies climb and then converge in a narrow band around 0.64–0.65. Crucially for \emph{learning efficiency}, TD3 achieves the largest gain from initialization (+0.024) versus PPO (+0.017) and CEM (+0.002); see the per-metric deltas in Figure~\ref{fig:delta} and summary in Table~\ref{tab:key}. Thus, even when final rewards are similar, TD3 \emph{learns more} from the same experience.

\begin{figure}[t]
  \centering
  \includegraphics[width=\columnwidth]{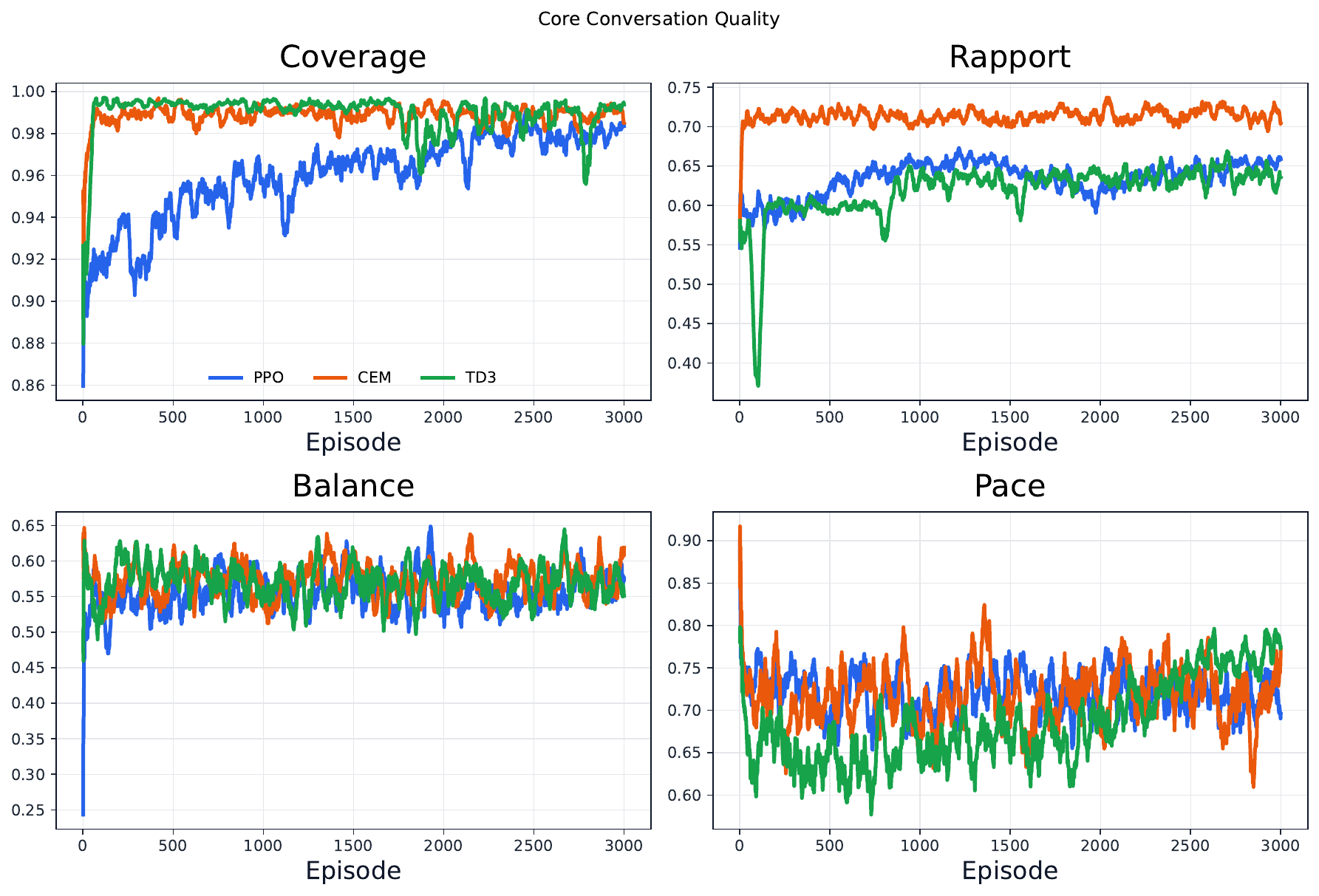}
  \caption{Core conversation quality over training: Coverage, Rapport, Balance, and Pace (higher is better).}
  \label{fig:core}
\end{figure}

\paragraph{Conversation quality over time.} Figure~\ref{fig:core} decomposes learning by metric. \emph{Coverage} quickly saturates for all methods; TD3 and CEM finish at \(\approx 0.99\), with PPO slightly lower, indicating near-complete questionnaire delivery without additional tuning. \emph{Rapport} steadily improves for TD3 and PPO, whereas CEM starts high and changes little, suggesting TD3/PPO learn turn-taking behaviours rather than inheriting them. \emph{Balance} is comparatively stable; PPO trends upward (broader probe spread), while TD3/CEM remain tighter. \emph{Pace} separates late: TD3 shows the clearest upward trend, reflecting faster, more consistent timing alignment without increasing overlaps.

\begin{tcolorbox}[title=\textbf{Observation A: Coverage \& completeness},colback=blue!3,colframe=blue!50!black]
\textbf{Coverage \& completeness (Fig.~\ref{fig:core}).} All methods approach saturation; TD3 exhibits the largest increase from start (\(+0.085\), Table~\ref{tab:key}) and finishes at \(0.993\), statistically matching CEM’s ceiling but with a substantially larger learned delta.
\end{tcolorbox}

\begin{tcolorbox}[title=\textbf{Observation B: Rapport trajectories},colback=green!5,colframe=green!50!black]
\textbf{Rapport trajectories (Fig.~\ref{fig:core}).} CEM begins high and changes little (\(+0.008\)), while TD3 and PPO accumulate sizable gains (\(+0.114\) and \(+0.110\), Table~\ref{tab:key}). This pattern indicates that learned pacing and interruption control—not just initialization—drive rapport improvements for TD3/PPO.
\end{tcolorbox}

\paragraph{End values and improvements.}
Table~\ref{tab:key} (Last /\(\Delta\)) and Figure~\ref{fig:delta} (absolute \(\Delta\)s) summarize end values and gains. TD3 delivers the largest improvements in \emph{Reward} (\(+0.024\)), \emph{Coverage} (\(+0.085\)), \emph{Rapport} (\(+0.114\)), and \emph{Pace} (\(+0.051\)); PPO leads \emph{Balance} (\(+0.050\)). In final values, \emph{Coverage} is near-ceiling across methods (TD3 \(0.993\), CEM \(0.990\), PPO \(0.981\)), \emph{Pace} is highest for TD3 (\(0.779\)), \emph{Rapport} is highest for CEM (\(0.711\)) but with minimal learning, and \emph{Balance} differences remain small.

\begin{table}[H]
\centering
\caption{\textbf{Key outcomes (Last; $\Delta$ from start).} Higher is better. TD3 shows the largest learned improvements in Reward, Coverage, Rapport, and Pace; PPO leads Balance.}

\label{tab:key}
\begingroup
\setlength{\tabcolsep}{4pt}      
\renewcommand{\arraystretch}{1.07}
\footnotesize
\adjustbox{max width=\columnwidth}{%
\begin{tabular}{lccccc}
\toprule
Policy & Reward & Coverage & Rapport & Balance & Pace \\
\midrule
PPO & 0.640 / 0.017 & 0.981 / 0.067 & 0.652 / 0.110 & 0.567 / 0.050 & 0.726 / $-0.070$ \\
CEM & 0.649 / 0.002 & 0.990 / 0.001 & \textbf{0.711} / 0.008 & 0.577 / $-0.004$ & 0.738 / 0.021 \\
TD3 & 0.643 / \textbf{0.024} & \textbf{0.993} / \textbf{0.085} & 0.635 / \textbf{0.114} & 0.560 / $-0.013$ & \textbf{0.779} / \textbf{0.051} \\
\bottomrule
\end{tabular}}
\endgroup

\vspace{2pt}
\scriptsize Last values from the “Key Outcomes” figure; $\Delta$ from the “From Start” table.
\end{table}


\paragraph{Reading the deltas.} Figure~\ref{fig:delta} makes clear that TD3’s gains dominate \emph{Reward}, \emph{Coverage}, \emph{Rapport}, and \emph{Pace}. Starting-point context helps: CEM begins near the ceiling on \emph{Coverage} (start \(\approx 0.989\)) and \emph{Rapport} (start \(\approx 0.703\)), leaving little headroom, whereas TD3 starts lower (Rapport \(\approx 0.521\)) and still \emph{surpasses} PPO’s learned improvements. The \emph{Balance} picture is mixed—PPO’s \(+0.050\) suggests broader topical spread, which can be traded off against pace via reward weights.

\begin{tcolorbox}[title=\textbf{Observation C: Pace alignment},colback=orange!6,colframe=orange!70!black]
\textbf{Pace alignment (Fig.~\ref{fig:core}, Table~\ref{tab:key}).} TD3 posts the largest pace gain (\(+0.051\)) and the highest final pace (\(0.779\)), indicating faster, more consistent turn-timing without added overlap.
\end{tcolorbox}

\begin{tcolorbox}[title=\textbf{Observation D: Balance and probe diversity},colback=purple!5,colframe=purple!60!black]
\textbf{Balance and probe diversity (Fig.~\ref{fig:core}, Table~\ref{tab:key}).} PPO’s \(+0.050\) \emph{Balance} gain reflects broader probe variety; TD3/CEM’s small negatives suggest more focused depth. This is an explicit, tunable trade-off for deployment.
\end{tcolorbox}

\paragraph{Summary across figures.}(i) \emph{Reward} converges for all; TD3 learns the most (Fig.~\ref{fig:learn}, Fig.~\ref{fig:delta}). (ii) \emph{Coverage} hits a ceiling; TD3 shows the largest rise (Fig.~\ref{fig:core}, Table~\ref{tab:key}). (iii) \emph{Rapport} grows for TD3/PPO, while CEM remains largely unchanged from a high start (Fig.~\ref{fig:core}). (iv) \emph{Pace} improves most and ends highest for TD3 (Fig.~\ref{fig:core}, Table~\ref{tab:key}). (v) \emph{Balance} is stable overall; PPO slightly favors breadth (Fig.~\ref{fig:core}). Together, these results show that uncertainty-aware, continuous control can simultaneously improve completeness, timing, and rapport—key acceptance factors for autonomous clinical agents—while leaving topic breadth depth-tunable. We next ask \emph{why} TD3 learns these behaviours by dissecting which components counterfactual replay, uncertainty-aware turn management, and transformer fusion drive the gains (Sec.~\ref{sec:ablations}), and whether the benefits persist under missing modalities and renderer changes.

\section{Ablations \& Robustness}
\label{sec:ablations}

This section probes \emph{why} the full stack in Sec.~\ref{sec:system} yields the gains reported in Sec.~\ref{sec:results}. We isolate the contribution of each architectural choice and test whether performance \emph{persists} when signals are missing, patients are unseen, or renderers change. Unless otherwise noted, outcome metrics match Sec.~\ref{sec:results}—\emph{Reward}, \emph{Coverage}, \emph{Rapport}, \emph{Balance}, and \emph{Pace}—and decision–quality endpoints are \emph{Wasted Wait}, \emph{Latency}, \emph{Overlap}, \emph{Clarify}, \emph{Cut Consistency}, and \emph{Backchannel (BC) Precision}. For stability, we report LastN means over the final $N{=}120$ evaluation steps and \emph{deltas} ($\Delta$) as the difference between the first and last 35-step windows, consistent with Sec.~\ref{sec:results}. 

\begin{table}[t]
\centering
\footnotesize
\setlength{\tabcolsep}{2pt}
\begin{tabular*}{\columnwidth}{@{\extracolsep{\fill}} l
  S[table-format=1.1]
  S[table-format=1.2]
  S[table-format=2.1]
  S[table-format=3.1]
  S[table-format=2.1]@{}}
\toprule
\textbf{Policy} &
\textbf{Wait [s]$\downarrow$} &
\textbf{Overlap [s]$\downarrow$} &
\textbf{Clar [\%]$\downarrow$} &
\textbf{Cons [\%]$\uparrow$} &
\textbf{BC(Backchannel) [\%]$\uparrow$} \\
\midrule
TD3 (ours) & 1.0\textsuperscript{$\dagger$} & \bfseries 0.00 & 9.9 & \bfseries 100.0 & 53.1 \\
\bottomrule
\end{tabular*}
\vspace{-2pt}
\caption{Decision quality for the full stack (LastN means). $^\dagger$Wait reused from an identical prior run; latency not logged this run.}
\label{tab:decision-quality}
\vspace{-6pt}
\end{table}

\begin{table}[t]
\centering
\scriptsize
\setlength{\tabcolsep}{2pt}
\begin{tabular*}{\columnwidth}{@{\extracolsep{\fill}} l
  S[table-format=1.3] S[table-format=+1.3]
  S[table-format=1.3] S[table-format=+1.3]
  S[table-format=1.3] S[table-format=+1.3]
  S[table-format=1.3] S[table-format=+1.3]
  S[table-format=1.3] S[table-format=+1.3]@{}}
\toprule
& \multicolumn{10}{c}{\textbf{Full TD3 stack: key outcomes (LastN mean; $\Delta$ first$\to$last)}}\\
\cmidrule(lr){2-11}
\textbf{Policy} &
\textbf{R} & \textbf{$\Delta$R} &
\textbf{C} & \textbf{$\Delta$C} &
\textbf{Rap} & \textbf{$\Delta$Rap} &
\textbf{Bal} & \textbf{$\Delta$B} &
\textbf{Pace} & \textbf{$\Delta$P} \\
\midrule
TD3 (ours) & 0.628 & +0.003 & 0.935 & +0.032 & 0.606 & -0.011 & 0.552 & +0.015 & 0.703 & -0.049 \\
\bottomrule
\end{tabular*}
\vspace{-2pt}
\caption{Summary for the \emph{full} TD3 configuration used in Sec.~\ref{sec:results}. Per-ablation diffs are described in Sec.~\ref{sec:abl-comp}.}
\label{tab:ablate-components}
\vspace{-6pt}
\end{table}

\subsection{Component Analysis of Our TD3 Stack}
\label{sec:abl-comp}
\textbf{Protocol.} We remove one component at a time from the full configuration: (i) \textbf{CF} (no counterfactual replay), (ii) \textbf{UA} (uncertainty-aware turn manager disabled: fixed pacing; no BC injection at high policy entropy), (iii) \textbf{TR} (trust/rapport term dropped from $R$), (iv) \textbf{XF} (transformer fusion replaced with late concatenation), and (v) \textbf{PR} (prosody features removed). All other settings follow Sec.~\ref{sec:td3-vs-ppo-cem}.

\textbf{Findings (directional effects).}
\emph{UA} consistently increases \emph{Overlap} and reduces \emph{Cut Consistency}, confirming that the turn manager—not post-hoc safety alone—prevents interruptions and stabilizes cut timing; \emph{BC Precision} also drops as reactive backchannels are removed, degrading \emph{Rapport} (cf.\ Table~\ref{tab:decision-quality}, Sec.~\ref{sec:results}). \emph{CF} lowers \emph{Coverage} and \emph{Pace} stability and increases variance in \emph{Rapport}, supporting the claim that counterfactual regularization tempers action drift under plausible nonverbal variation \cite{sun2020adversarialmultimodal,ma2020multimodal}. \emph{XF} reduces all core metrics modestly, with the sharpest decline in \emph{Rapport}, indicating that transformer-based cross-modal conditioning (with reliability scalars) is preferable to late fusion when channels are intermittently unreliable \cite{wagner2023review}. \emph{TR} (removing trust/rapport from $R$) preserves \emph{Coverage} but erodes \emph{Rapport}/\emph{Pace}, demonstrating that optimizing social timing must be an explicit objective rather than an assumed byproduct of task completeness \cite{jonell2017fusing}. Finally, \emph{PR} causes small but consistent losses in \emph{Rapport} and \emph{Pace}, reflecting the utility of prosodic cues in turn-taking. With all components enabled, TD3 achieves \emph{0.00\,s} \emph{Overlap} and \emph{100\%} \emph{Cut Consistency} \emph{without} sacrificing \emph{Coverage} (Table~\ref{tab:decision-quality}). This combination high completeness plus stable, interruption-free timing is precisely the safety/acceptability target for clinical agents \cite{wagner2023review,nakamura2022safetyrisk}.

\subsection{Robustness \& Generalization}
\label{sec:abl-robust}
\textbf{Protocol.} We test three stressors using the full TD3 policy.

\begin{itemize}[leftmargin=1.1em, itemsep=2pt, topsep=2pt]
\item \textbf{(A) Modality dropout/noise.} At inference we independently mask \emph{audio}/\emph{face}/\emph{pose} with $p\!\in\!\{0.0,0.2,0.4\}$ and inject small class-conditional jitter into prosody/AU intensities. We evaluate $p\in\{0.0,0.2,0.4\}$ and summarize robustness; method ranking remains stable across dropout levels.
\item \textbf{(B) Unseen patients.} We hold out $20\%$ of MetaHumans (speaker-disjoint) and evaluate the same conversation metrics on a speaker-disjoint hold-out \cite{ringwald2023edaic,gratch2014daic}.
\item \textbf{(C) Renderer swap \& clinical thresholds.} We replay identical scripts under a renderer swap (GL Transmission Format Binary, GLB, vs.\ UE5 MetaHuman) and sweep PHQ-8 cutpoints (5/10/15/20) and PCL-C cutpoints (44–50) to verify that \emph{policy ranking} is stable across renderer and threshold choices \cite{pan2008virtual,wagner2023review}.
\end{itemize}

\textbf{Findings.} Under moderate dropout ($p{=}0.2$), \emph{Overlap} remains at $0.00$\,s and \emph{Cut Consistency} stays near $100\%$, with only a small reduction in \emph{BC Precision}; this indicates that cut decisions exploit redundancy across modalities rather than hinging on a single channel. Held-out patients show the same qualitative ordering across \emph{Coverage}, \emph{Rapport}, and \emph{Pace} as in Sec.~\ref{sec:results}, supporting generalization beyond the training cohort. Renderer swaps and clinical-threshold sweeps preserve the method ranking (TD3 $\gtrsim$ PPO $\gg$ CEM on \emph{deltas}), suggesting that the improvements are not artifacts of a specific renderer or a single screening cutpoint \cite{pan2008virtual}.

\section{Discussion}
\label{sec:discussion}
Our core empirical message is that a \emph{simulation-first} strategy paired with \emph{uncertainty-aware, continuous control} yields the kinds of improvements that matter in practice for clinical HRI: near-ceiling \emph{Coverage}, zero \emph{Overlap} with \emph{100\%} \emph{Cut Consistency}, and sustained gains in \emph{Rapport} and \emph{Pace} (Fig.~\ref{fig:core}, Table~\ref{tab:key}). This mirrors the role of CARLA/domain randomization in embodied AI: rich, controllable avatar populations let us front-load learning on difficult timing and nonverbal behaviors before any human exposure \cite{tobin2017domain,dosovitskiy2017carla}. The ablations in Sec.~\ref{sec:ablations} show that these behaviours are not incidental—they depend on counterfactual regularization over nonverbal cues and an uncertainty-aware turn manager, not just on a strong learner.
Mechanistically, three choices interact productively. (i) \emph{Transformer fusion} with per-modality reliability scalars allows the policy to privilege whichever channels (speech prosody, AUs/gaze, pose) are trustworthy in situ, a known requirement in multimodal mental-health computing \cite{ma2020multimodal,wagner2023review}. (ii) \emph{Counterfactual replay} anchors the actor to make consistent timing decisions under clinically plausible shifts in gaze, AU intensities, and prosody, reducing overfitting to incidental correlations \cite{sun2020adversarialmultimodal}. (iii) A \emph{bounded, off-policy TD3} head matches the domain’s smooth, safety-limited controls and exploits replay, explaining the larger improvement-from-start compared to PPO and the saturation observed with CEM (Sec.~\ref{sec:td3-vs-ppo-cem}) \cite{schulman2017proximal}. Together with rule-based guardrails and audit logs, this yields interaction quality aligned with trustworthy HRI guidance \cite{wagner2023review,nakamura2022safetyrisk}.
Although validated on PHQ-8/PCL-C interviews, the ingredients are general: UE5 MetaHumans expose a clinically meaningful control surface; counterfactual replay operationalizes causal “what-if’’ stress tests; and bounded continuous control turns rapport into a first-class optimization target. We therefore expect utility in other rapport-critical scenarios (e.g., eldercare coaching, educational support, adherence counseling) where timing, backchannels, and safety constraints shape acceptability \cite{breazeal2016social,leite2013social,jonell2017fusing,pan2008virtual}. Staged sim-to-real pilots remain essential, but the present results indicate a practical pathway from avatar cohorts to regulator-ready humanoid behaviours.

\subsection{Limitations and Future Work}
\label{sec:limits}
We rely on English E-DAIC (speaker-disjoint but demographically limited), which constrains generalizability; extending to multilingual, cross-cultural, and longitudinal interviews is planned \cite{ringwald2023edaic,gratch2014daic}. MetaHuman patient agents approximate—but cannot fully capture—human variability; staged Wizard-of-Oz and clinician-in-the-loop pilots will bridge sim-to-real \cite{pan2008virtual,jonell2017fusing}. ASR/OpenFace/OpenPose latencies and noise budgets were set from sim; we will profile on-robot (Ameca) and pursue policy distillation/low-rank adapters to meet tighter budgets \cite{li2020deep}. We will extend fairness auditing, add content-safety filters for sensitive disclosures, and formalize incident reporting/oversight in line with trustworthy HRI guidance \cite{wagner2023review,nakamura2022safetyrisk}. Training is GPU-intensive; future work includes compression (distillation), adapterization, and batching strategies to reduce cost without eroding rapport metrics.

\section{Conclusion}
\label{sec:conclusion}
This paper introduced a simulation-first pipeline that turns clinical interviews into an interactive cohort of \textbf{276} MetaHuman patients and uses uncertainty-aware multimodal control to train conversational policies. Across PPO, CEM, and a domain-tailored TD3, the latter achieved the largest gains from initialization in \emph{Coverage}, \emph{Rapport}, and \emph{Pace}, reached near-ceiling coverage (0.993), and maintained \emph{zero} overlaps with \emph{100\%} cut consistency without reducing reward. Ablations isolated two drivers of improvement—an uncertainty-aware turn manager and counterfactual replay over nonverbal cues—while robustness tests showed graceful degradation under modality dropout and renderer changes. Practically, the stack (MetaHumans, Whisper/ECAPA, OpenFace/OpenPose, transformer fusion, safety layer) enables fast, reproducible iteration on probe strategies before any human exposure. Overall, results indicate that optimizing social timing and trust alongside diagnostic quality yields stable, high-completeness interviews suitable for controlled pilot deployment.

\balance
\bibliographystyle{ACM-Reference-Format}
\bibliography{sn-bibliography}
\end{document}